\documentclass[journal]{IEEEtran}
\usepackage{amsmath,amsfonts}
\usepackage{algorithmic}
\usepackage{algorithm}
\usepackage{array}
\usepackage[caption=false,font=normalsize,labelfont=sf,textfont=sf]{subfig}
\usepackage{textcomp}
\usepackage{stfloats}
\usepackage{url}
\usepackage{verbatim}
\usepackage{graphicx}
\usepackage{cite}
\usepackage{multirow}
\usepackage{booktabs}
\usepackage{mathptmx}
\begin{document}

\title{SemAgent: Semantic-Driven Agentic AI Empowered Trajectory Prediction in Vehicular Networks}

\author{Lin Zhu, Kezhi Wang,~\IEEEmembership{Senior Member,~IEEE}, Luping Xiang,~\IEEEmembership{Senior Member,~IEEE}, and Kun Yang,~\IEEEmembership{Fellow,~IEEE}
\thanks{Lin Zhu (zhulina@smail.nju.edu.cn), Luping Xiang (luping.xiang@nju.edu.cn), and Kun Yang (kunyang@essex.ac.uk) are with the State Key Laboratory of Novel Software Technology, Nanjing University, Nanjing, China, and the School of Intelligent Software and Engineering, Nanjing University (Suzhou Campus), Suzhou, China.}
\thanks{Kezhi Wang (Kezhi.Wang@brunel.ac.uk) is with the Department of Computer Science, Brunel University London, UK.}}

\markboth{Journal of \LaTeX\ Class Files,~Vol.~14, No.~8, August~2021}%
{Shell \MakeLowercase{\textit{et al.}}: A Sample Article Using IEEEtran.cls for IEEE Journals}


\maketitle

\begin{abstract}
Efficient information exchange and reliable contextual reasoning are essential for vehicle-to-everything (V2X) networks. Conventional communication schemes often incur significant transmission overhead and latency, while existing trajectory prediction models generally lack environmental perception and logical inference capabilities. This paper presents a trajectory prediction framework that integrates semantic communication with Agentic AI to enhance predictive performance in vehicular environments. In vehicle-to-infrastructure (V2I) communication, a feature-extraction agent at the Roadside Unit (RSU) derives compact representations from historical vehicle trajectories, followed by semantic reasoning performed by a semantic-analysis agent. The RSU then transmits both feature representations and semantic insights to the target vehicle via semantic communication, enabling the vehicle to predict future trajectories by combining received semantics with its own historical data. In vehicle-to-vehicle (V2V) communication, each vehicle performs local feature extraction and semantic analysis while receiving predicted trajectories from neighboring vehicles, and jointly utilizes this information for its own trajectory prediction. Extensive experiments across diverse communication conditions demonstrate that the proposed method significantly outperforms baseline schemes, achieving up to a 47.5\% improvement in prediction accuracy under low signal-to-noise ratio (SNR) conditions.
\end{abstract}

\begin{IEEEkeywords}
Semantic communication, Large Language Models(LLMs), Autonomous Vehicles, Trajectory prediction.
\end{IEEEkeywords}

\section{Introduction}
\IEEEPARstart{W}{ith} the rapid evolution of 5G and emerging 6G wireless technologies, vehicle-to-everything (V2X) \cite{clancy2024wireless} systems have experienced significant advancements. V2X enables real-time information exchange among vehicles, infrastructure, pedestrians, and cloud services \cite{kumari2024investigating}, and has become a fundamental enabler for intelligent transportation and autonomous driving. Existing V2X communication frameworks are primarily grounded in Shannon’s information theory, which focuses on ensuring accurate bit-level reproduction at the receiver. To mitigate channel impairments, redundancy is incorporated for error control \cite{luo2022semantic}, and the encoded bitstream is subsequently modulated and transmitted over the wireless channel. Upon reception, the receiver demodulates and decodes the signal to reconstruct the original bit sequence. While effective for bit-accurate transmission, this paradigm inherently overlooks the semantic utility of the transmitted data \cite{xin2024semantic}. Moreover, raw environmental data—such as high-resolution sensor streams—demands substantial transmission and processing resources, which contradicts the limited computational capabilities of on-board units and the dynamically varying nature of vehicular environments.

To address these limitations, semantic communication has emerged as a promising paradigm. Rather than transmitting raw data, semantic communication \cite{shao2024theory} emphasizes the conveyance of task-relevant meaning—i.e., the specific information needed for inference or decision-making. By jointly leveraging semantic extraction and contextual understanding, it enables efficient transmission of key driving-related information, such as motion intent, thereby reducing computational overhead while enhancing situational awareness \cite{sun2025edge}. In parallel, most existing trajectory prediction approaches \cite{jia2023towards, geng2023dynamic, geng2023physics} concentrate on optimizing deep neural models to capture the spatiotemporal characteristics of vehicle trajectories. However, these methods often lack comprehensive modeling of environmental context and dynamic interactions. Real-world vehicle motion is inherently influenced not only by individual behavior but also by neighboring agents, roadway conditions, and real-time communication with roadside infrastructure \cite{gimenez2024semantic}. Motivated by this gap, this work introduces Agentic AI into V2I \cite{sultana2024toward} and V2V \cite{anizan2025vehicle} trajectory prediction to enhance vehicles’ interpretive and reasoning capabilities in complex environments.

Agentic AI \cite{murugesan2025rise} refers to AI systems equipped with autonomous perception, goal-driven reasoning, decision-making, and adaptive behavior. Unlike conventional rule-based systems, Agentic AI exhibits deep contextual reasoning, self-directed planning, and robust adaptability, enabling it to operate effectively in dynamic and uncertain environments \cite{masterman2024landscape}. In V2I and V2V semantic communication scenarios, systems must account for uncertainties such as fluctuating traffic flow, weather conditions, incidents, and road closures. Agentic AI leverages real-time multi-source information \cite{gu2024survey} exchanged between vehicles and RSUs (V2I) and across vehicles (V2V) to perform fine-grained inference of future vehicle trajectories. By integrating such contextual semantics, Agentic AI can more accurately predict traffic trends, road hazards, and vehicle behaviors, thereby improving trajectory prediction in dynamic environments and supporting safer and more efficient path planning.

Building on these observations, this paper proposes a trajectory prediction scheme that fuses semantic communication with Agentic AI. The objective is to exploit the reasoning capabilities of Agentic AI and the compact, task-oriented transmission enabled by semantic communication to support accurate trajectory forecasting in complex traffic environments. The main contributions of this work are summarized as follows:
\begin{itemize}
\item We introduce a multi-agent collaborative trajectory prediction framework based on Agentic AI, consisting of a feature-extraction agent, a semantic-analysis agent, and a trajectory-prediction agent. The feature-extraction agent processes the raw data, the semantic-analysis agent performs semantic reasoning on extracted features, and the trajectory-prediction agent fuses these results with historical trajectories to estimate future vehicle motion.
\item We develop a multi-communication-mode trajectory prediction strategy. In V2I scenarios, RSUs utilize feature-extraction and semantic-analysis agents to derive high-level semantic representations of surrounding traffic, which assist vehicles in predicting future trajectories. In V2V scenarios, vehicles exchange predicted future trajectories to collaboratively enhance prediction robustness. To eliminate redundant transmissions and improve semantic coherence, both communication modes employ semantic communication.
\item We validate the proposed scheme using the US-101 subset of the NGSIM dataset. Experimental results show that the proposed approach outperforms baseline methods in both Average Displacement Error (ADE) and Final Displacement Error (FDE), achieving up to a 47.5\% improvement in prediction accuracy under low signal-to-noise ratio (SNR) conditions.
\end{itemize}

The remainder of this paper is organized as follows. Section~\ref{sec:2} reviews related work. Section~\ref{sec:3} introduces the system model. Section~\ref{sec:4} describes the proposed V2I trajectory prediction scheme. Section~\ref{sec:5} details the V2V trajectory prediction process. Section~\ref{sec:6} presents the experimental evaluation and results. Section~\ref{sec:7} concludes the paper.

\section{Related Work}
\label{sec:2}
Semantic communication has been increasingly adopted to improve transmission efficiency and data interpretation in autonomous driving. Ribouh \textit{et al.} \cite{ribouh2024seecad} proposed SEECAD, a semantic end-to-end communication framework for image transmission that utilizes a shared knowledge base. Feng \textit{et al.} \cite{feng2024semantic} designed a unified multi-user semantic communication architecture for multimodal data and multi-task scenarios, enabling cooperative perception and decision-making. Lv \textit{et al.} \cite{lv2024importance} developed a semantic communication system for vehicular image segmentation using a Swin Transformer-based encoder-decoder with importance-aware perceptual loss and multi-scale extraction to improve object segmentation accuracy.

In trajectory prediction, deep learning remains the dominant methodology \cite{shiwakoti2023deep, yin2025deep, ding2023incorporating}. Qin \textit{et al.} \cite{qin2023spatiotemporal} introduced a hierarchical CapsNet-based architecture for spatiotemporal trajectory modeling. To capture vehicle interactions, Shi \textit{et al.} \cite{shi2022integrated} developed an integrated 2-D trajectory prediction model employing attention mechanisms, Bi-LSTM, and temporal CNNs. Shen \textit{et al.} \cite{shen2023spatio} proposed a spatiotemporal interactive graph convolutional network incorporating spatial autocorrelation priors and gated recurrent units to dynamically model inter-vehicle influences.

Recent works \cite{ferrag2025llm, costarelli2024gamebench, wu2025agentic} highlight the strong reasoning capabilities of Agentic AI. Grötschla \textit{et al.} \cite{grotschla2025agentsnet} proposed AgentsNet to evaluate collaborative problem-solving in multi-agent networks, showing that modern large language models exhibit promising emergent behaviors. Luo \textit{et al.} \cite{luo2025oneke} presented OneKE, a multi-agent framework with a configurable knowledge base for structured extraction. Kannan \textit{et al.} \cite{kannan2024smart} demonstrated the use of LLMs for multi-robot task planning, translating high-level instructions into executable multi-robot action plans.

Despite promising progress, current Agentic-AI-based approaches for multi-agent trajectory prediction still struggle with unified semantic scene interpretation and joint inference across agents. Existing methods primarily enhance single-agent interpretability rather than addressing collective spatiotemporal reasoning. To bridge this gap, we propose a semantic-communication-enhanced trajectory prediction scheme powered by Agentic AI that enables collaborative, context-aware trajectory forecasting among multiple vehicles in complex traffic environments.

\section{System Model}
\label{sec:3}
Vehicle trajectory prediction is a fundamental component of autonomous driving systems, particularly in highly dynamic and complex traffic environments where accurate forecasting significantly enhances safety and decision-making performance. The task aims to infer future vehicle motion patterns from historical observations. As formulated in \cite{deo2018convolutional}, trajectory prediction can be regarded as estimating the probability distribution of future trajectories conditioned on known past trajectories. This inherently requires the ability to interpret historical motion data and reason about future behavioral tendencies.

Trajectory prediction constitutes a cognitively demanding process that integrates perception, reasoning, and decision-making. Although traditional end-to-end neural models can map historical trajectories to future predictions, they generally lack interpretability, controllability, and deep task-level understanding. In contrast, the Agentic AI framework adopted in this study provides autonomous perception, contextual reasoning, and adaptive decision-making capabilities. It enables vehicles to interpret complex environments, integrate multi-source information, and generate more rational and accurate predictions. This paper investigates trajectory prediction under both V2I and V2V communication scenarios, as detailed below.

\subsection{V2I Communication Model}
\begin{figure}[htbp]       
     \centering
        \includegraphics[width=\linewidth]{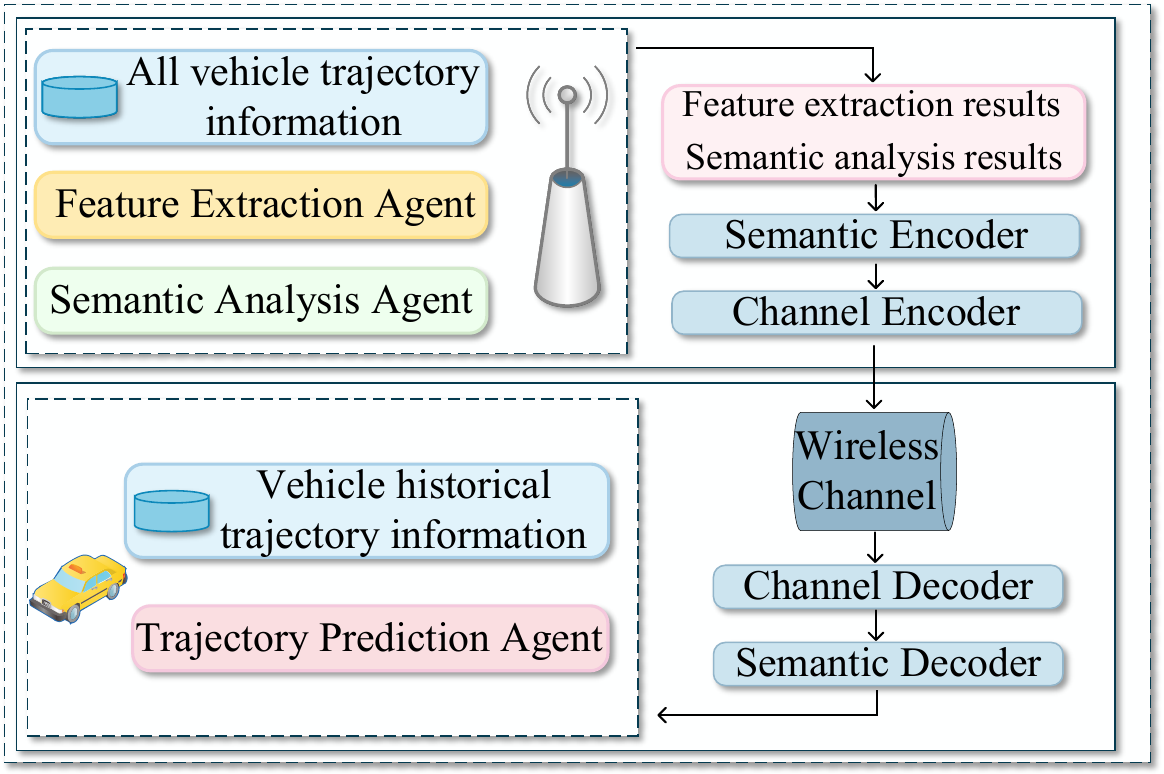}
        \caption{V2I communication framework.}
        \label{fig:mt1}
\end{figure}

The V2I communication framework is illustrated in Fig.~\ref{fig:mt1}. Let vehicle be denoted by $v$. Each RSU maintains information about all vehicles within its coverage range and hosts a feature-extraction agent and a semantic-analysis agent. Information exchange between RSU and vehicle $v$ is realized through semantic communication, wherein the RSU transmits the outputs of these two agents to vehicle $v$. Vehicle $v$ further combines its own historical data with the received semantic and feature representations using a trajectory prediction agent. The details of all agent modules are presented in Section~\ref{sec:4}.

The historical trajectory of vehicle $v$ over the past interval $t_T$ is denoted as

\begin{equation}
\mathcal{P}_v (t_T)=\{P_v(t_1),P_v(t_2),\dots,P_v(t_T)\}, \label{eq:eq1}
\end{equation}
where each position point $P_v(t_i)$ is represented as
\begin{equation}
P_v(t_i)=\{p_v^1(t_i),p_v^2(t_i),\dots,p_v^N(t_i)\},\label{eq:eq2}
\end{equation}
with $p_v^n(t_i)=(x_v^n(t_i),y_v^n(t_i))$ denoting the two-dimensional coordinates of vehicle $v$ at time $t_i$.

\begin{figure}[htbp]       
     \centering
        \includegraphics[width=\linewidth]{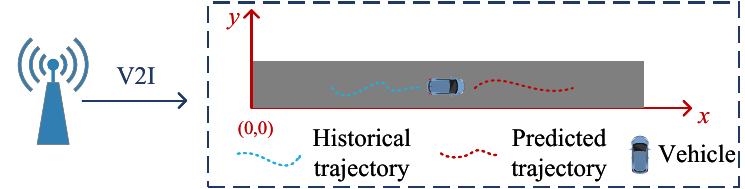}
        \caption{Illustration of the V2I prediction scenario.}
        \label{fig:mt2}
\end{figure}

Fig.~\ref{fig:mt2} depicts the prediction scenario for V2I. The objective is to estimate the trajectory of vehicle $v$ over a future horizon $S$ by integrating its historical trajectory and traffic semantics from the RSU. The predicted trajectory is expressed as
\begin{equation}
\mathcal{\hat{P}}_v (t_{T+S})=\{\hat{P}_v(t_{T+1}),\hat{P}_v (t_{T+2}),\dots,\hat{P}_v (t_{T+S})\}, \label{eq:eq3}
\end{equation}
where, each predicted position $\hat{P}_v (t_{T+s})$ is represented by

\begin{equation}
\hat{P}_v (t_{T+s})=\{\hat{p}_v^1(t_{T+s}),\hat{p}_v^2 (t_{T+s}),\dots,\hat{p}_v^N (t_{T+s})\}, \label{eq:eq4}
\end{equation}
with $\hat{p}_v^n (t_{T+s}) = (\hat{x} _v^n(t_{T+s}), \hat{y}_v^n(t_{T+s}))$ denoting the predicted coordinates. The algorithmic solution is presented in Section~\ref{sec:4}.

\subsection{V2V Communication Model}
\begin{figure}[htbp]       
     \centering
        \includegraphics[width=\linewidth]{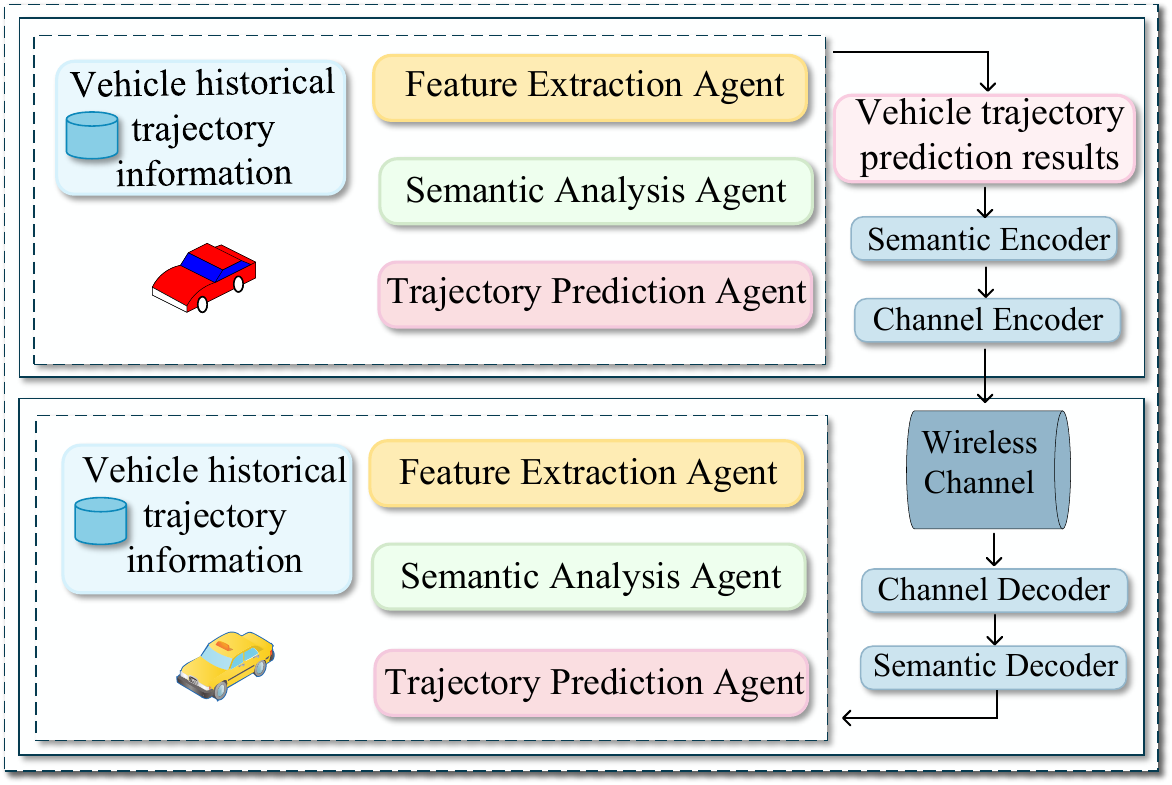}
        \caption{V2V communication framework.}
        \label{fig:mt3}
\end{figure}

The V2V communication framework is depicted in Fig.~\ref{fig:mt3}. Let the vehicle set be $V=\{v_1,v_2,\dots,v_n\}$. Vehicles exchange information with neighboring vehicles through semantic communication; thus, vehicles $v_i$ and $v_k$ communicate by transmitting semantically encoded information. Each vehicle $v_i$ maintains its own historical data and receives the predicted future trajectories of neighboring vehicles $v_k$. The V2V framework consists of a feature-extraction agent, a semantic-analysis agent, and a trajectory prediction agent.

The historical trajectory of vehicle $v_i$ over the interval $t_T$ is defined as
\begin{equation}
\mathcal{P}_{v_i} (t_{T})=\{P_{v_i}(t_{1}),P_{v_i} (t_{2}),\dots,P_{v_i} (t_{T})\}, \label{eq:eq5}
\end{equation}
where 
\begin{equation}
P_{v_i} (t_i) = \{p_{v_i}^1 (t_i), p_{v_i}^2 (t_i),\dots, p_{v_i}^N (t_i)\}, \label{eq:eq6}
\end{equation}
 and $p_{v_i}^n (t_i) = (x_{v_i}^n (t_i), y_{v_i}^n (t_i))$ indicates the two-dimensional position of vehicle $v_i$ at time $t_i$.

\begin{figure}[htbp]       
     \centering
        \includegraphics[width=\linewidth]{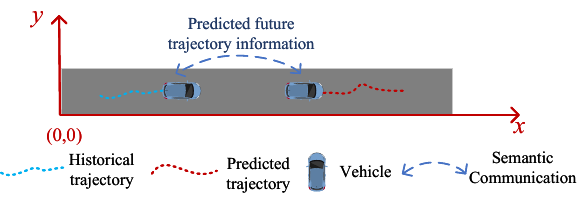}
        \caption{Illustration of the V2V prediction scenario.}
        \label{fig:mt4}
\end{figure}

The prediction scenario for V2V is shown in Fig.~\ref{fig:mt4}. Vehicle $v_i$ communicates with neighboring vehicles $v_k$ to obtain their predicted future trajectories through semantic communication. The objective is to estimate the future trajectory of $v_i$ over horizon $S$ using both its own historical trajectory and the received predictions from neighboring vehicles:
\begin{equation}
\mathcal{{\hat{P}}}_{v_i} (t_{T+S}) = \{\hat{P}_{v_i} (t_{T+1}), \hat{P}_{v_i} (t_{T+2}), \dots, \hat{P}_{v_i} (t_{T+S})\}, \label{eq:eq7}
\end{equation}
where
\begin{equation}
\hat{P}_{v_i} (t_{T+s}) = \{\hat{P}_{v_i}^1 (t_v), \hat{P}_{v_i}^2 (t_{T+S}), \dots, \hat{P}_{v_i}^N (t_{T+S})\}, \label{eq:eq8}
\end{equation}
with ${\hat{p}}_{v_i}^n (t_{T+s}) = (\hat{x}_{v_i}^n (t_{T+s}), \hat{y}_{v_i}^n (t_{T+s}))$ denoting the predicted coordinates. The complete V2V trajectory prediction mechanism is described in Section~\ref{sec:5}.

\section{Semantic-Driven Agentic AI Empowered Framework with RSU } \label{sec:4}
In the V2I scenario, this paper proposes a semantic-driven Agentic AI empowered trajectory prediction framework that jointly exploits historical vehicle motion data and traffic environment information provided by RSUs to forecast vehicle trajectories over a future horizon of $T_f$ time steps. Considering the limited computational resources and data interpretation capabilities of on-board units, semantic communication is introduced to reduce transmission overhead while enhancing the vehicles' understanding of the surrounding environment. Within this framework, vehicles and RSUs exchange semantically compressed information rather than raw data streams, thereby improving communication efficiency and task relevance.

\begin{figure*}[htbp]       
     \centering
        \includegraphics[width=\textwidth]{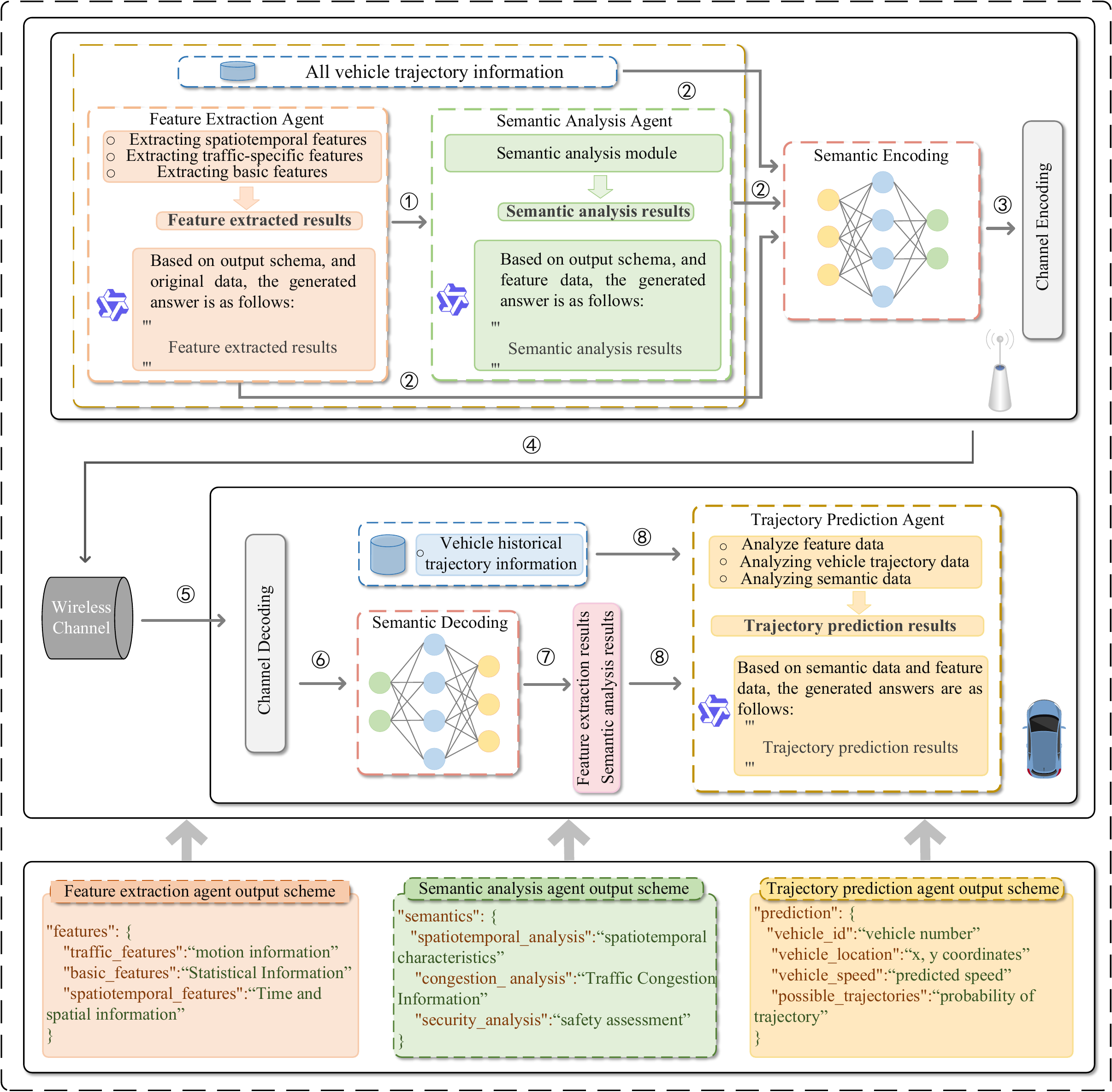}
        \caption{V2I network diagram. The RSU transmits trajectory-related information within its coverage via semantic communication. The target vehicle performs feature extraction and semantic analysis and then predicts its future trajectory by integrating its own historical trajectory data.}
        \label{fig:mt5}
\end{figure*}

The proposed architecture comprises a feature extraction agent, a semantic analysis agent, a trajectory prediction agent, a semantic encoder–decoder pair, and a channel encoder–decoder pair. The overall network structure is shown in Fig.~\ref{fig:mt5}. On the RSU side, the system maintains traffic environment information within the RSU coverage, and deploys a feature extraction agent, a semantic analysis agent, a semantic encoder, and a channel encoder. On the vehicle side, the system includes local vehicle information, a channel decoder, a semantic decoder, and a trajectory prediction agent. Feature extraction results and semantic analysis outputs are transmitted from the RSU to the vehicle via the semantic communication pipeline.

\subsection{Feature Extraction Agent in V2I}

Since the surrounding traffic environment has a direct impact on trajectory prediction accuracy, this work first performs feature extraction on traffic environment information within the RSU coverage, thereby providing auxiliary information for subsequent semantic analysis and trajectory prediction. The primary goal of the feature extraction agent is to derive compact, informative representations of global traffic states, road attributes, and vehicle behaviors from high-dimensional environment data.

Operationally, the feature extraction agent first ingests global traffic environment data and computes motion-related features (e.g., speed and acceleration), spatiotemporal descriptors (e.g., vehicle positions over time), and statistical indicators (e.g., mean and standard deviation) that characterize data distributions. To ensure consistency and structure, feature extraction is carried out following a predefined schema. The resulting feature set captures multiple dimensions, including temporal, spatial, and traffic-flow aspects, thereby encompassing factors that may influence vehicle trajectories. These features are then forwarded to the semantic analysis agent and the trajectory prediction agent, enhancing both the fidelity of semantic understanding and the accuracy of the final trajectory predictions.

\subsection{Semantic Analysis Agent in V2I}

To enable vehicles to fully comprehend and interpret the surrounding traffic environment, a semantic analysis agent is further employed. Its objective is to improve the robustness and accuracy of trajectory prediction by performing deep semantic parsing of the multi-source traffic features produced by the feature extraction agent.

Using the environmental feature data, the semantic analysis agent executes multidimensional situation assessment and semantic modeling and produces a structured semantic report. This report encompasses anomaly detection (e.g., accidents, temporary road closures, and sudden congestion), congestion and temporal analysis (e.g., traffic flow evolution across time periods and road segments), as well as spatial and trajectory pattern analysis (e.g., lane-change behaviors, intersection traversal patterns, and speed–space distributions). Through these analyses, the agent uncovers latent traffic regularities and interaction patterns that are not directly observable from raw data. By embedding semantic-level information into the trajectory prediction process, the proposed framework equips intelligent connected vehicles with enhanced environmental understanding, thereby supporting more informed and reliable trajectory forecasting.

\subsection{Semantic Communication}

In the proposed V2I system, information exchange between RSUs and vehicles is realized via semantic communication. At the RSU, the feature extraction agent first derives feature vectors from global traffic environment data, and the semantic analysis agent performs contextual reasoning based on these features. The resulting feature extraction outputs and semantic analysis outputs are then processed by a semantic encoder, followed by a channel encoder, and transmitted over the wireless channel to the target vehicle.

At the receiver, the vehicle uses a channel decoder to mitigate channel impairments, followed by a semantic decoder to reconstruct semantically meaningful representations of the global environment state. These reconstructed semantic features provide high-level auxiliary information for the downstream trajectory prediction task. In what follows, the feature extraction result is denoted by $fr$, and the semantic analysis result is denoted by $sr$.

\begin{figure}[htbp]       
     \centering
        \includegraphics[width=\linewidth]{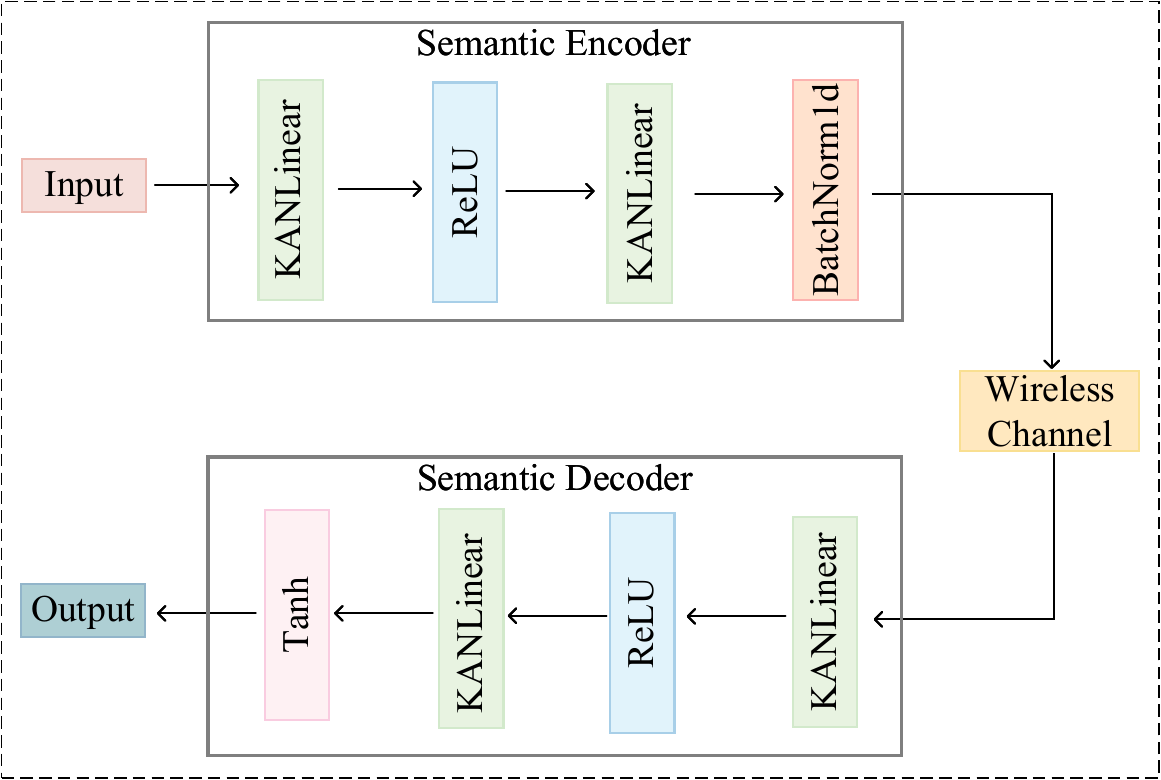}
        \caption{Network architecture of the semantic communication system.}
        \label{fig:mt20}
\end{figure}

The network architecture of the semantic communication module is shown in Fig.~\ref{fig:mt20}. The system follows an encoder–channel–decoder paradigm. On the transmitter side, input features are first processed by two KANLinear layers. The first KANLinear layer (384$\to$128) performs nonlinear compression and semantic refinement of high-dimensional inputs, allowing the model to extract discriminative high-level semantic features. A ReLU activation function is employed to suppress noise and enhance sparsity in the learned representations. The second KANLinear layer (128$\to$32) further compresses the feature dimension to adapt to the physical-layer bandwidth constraint, effectively acting as a semantic compression code. In conjunction with a BatchNorm1d layer, this structure stabilizes gradient propagation and mitigates overfitting during training, thereby improving robustness and generalization.

The encoded semantic features are transmitted through a wireless channel module that emulates channel fading and additive noise, allowing evaluation under realistic non-ideal communication conditions. On the receiver side, a symmetric decoding architecture is adopted. A KANLinear(32$\to$128) layer first expands the compressed feature dimension and initiates structural restoration in the semantic space, followed by a ReLU activation to enhance feature discrimination. Finally, a KANLinear(128$\to$384) layer completes semantic reconstruction, while a Tanh activation constrains the output range, ensuring that decoded semantic features remain physically interpretable.

\begin{itemize}
    \item[(1)] Semantic Encoding: To reduce semantic distortion during wireless transmission, semantic encoding is performed through the semantic encoder $F_{se}(\cdot)$. The semantic encoding process can be expressed as 
    \begin{equation}
    s_{fr} = F_{se}(x_{fr}, c_{fr}, \alpha), \label{eq:eq9}
    \end{equation}
    \begin{equation}
    s_{sr} = F_{se}(x_{sr}, c_{sr}, \alpha), \label{eq:eq10}
    \end{equation}

    \noindent where, $x_{fr}$, $x_{sr}$ are the original vectors emitted, $s_{fr}$ and $s_{sr}$ represent semantic encoding, $F_{se}(\cdot)$ is a neural network. The $k$-dimensional vector $s_{fr}$ output from $F_{se}(\cdot)$ represents the semantic information of the feature extraction results, while the $k$-dimensional vector $s_{sr}$ represents the semantic information of the semantic analysis results. $\alpha$ denotes the parameter of the semantic encoder $F_{se}(\cdot)$. $c_{fr}$ and $c_{sr}$ denote the channel bandwidth ratios, representing the ratio of the feature vector dimensions $k$ for the original data dimensions $m_{fr}$ and $m_{sr}$, respectively. This can be expressed as: 
    \begin{equation}
    c_{fr} = \frac{k}{m_{fr}}, \label{eq:eq11}
    \end{equation}
    \begin{equation}
    c_{sr} = \frac{k}{m_{sr}}. \label{eq:eq12}
    \end{equation}

    \item[(2)]  Wireless Communication: When transmitting over a fading wireless channel, the vector $s_{fr}$ and $s_{sr}$ are subject to transmission losses including distortion and noise. This transmission process can be modeled as:
    \begin{equation}
    y_{fr} = H \cdot s_{fr} + N, \label{eq:eq13}
    \end{equation}
    \begin{equation}
    y_{sr} = H \cdot s_{sr} + N, \label{eq:eq14}
    \end{equation}

    \noindent where $y_{fr}$ and $y_{sr}$ denote the received complex vectors, $H$ represents the channel gain between the transmitter and receiver, and $N$ denotes additive white Gaussian noise (AWGN).  

    \item[(3)] Semantic Decoding: Upon receiving vectors $y_{fr}$ and $y_{sr}$, the semantic decoder reconstructs them. This process can be represented as follows: 
    \begin{equation}
    \hat{x}_{fr} = F_{sd}(y_{fr}, c_{fr}, \beta),  \label{eq:eq15}
    \end{equation}
    \begin{equation}
    \hat{x}_{sr} = F_{sd}(y_{sr}, c_{sr}, \beta),  \label{eq:eq16}
    \end{equation}

    \noindent where, $F_{sd} (\cdot)$ is a neural network. The vectors $y_{fr}$ and $y_{sr}$ output from $F_{sd} (\cdot)$ represent the reconstructed feature extraction results and semantic analysis results after decoding, respectively. $\beta$ denotes the parameters of the semantic decoder $F_{sd} (\cdot)$.
    
\end{itemize}

\subsection{Trajectory Prediction Agent in V2I}
To further improve prediction accuracy, the trajectory prediction agent fuses historical vehicle trajectories, the reconstructed feature extraction results, and semantic analysis information. First, the agent loads and preprocesses the semantically transmitted environmental features and semantic descriptors. These are then integrated with the vehicle’s own historical trajectory data to form an enriched input representation that captures both local motion characteristics and global traffic context.

This enhanced representation incorporates information such as traffic-flow variations, congestion levels, and spatiotemporal fluctuations received through semantic communication. The trajectory prediction agent applies deep learning and reasoning to this combined input, identifying underlying motion patterns and traffic regularities. On this basis, it generates short-term predictions of the vehicle’s future trajectory over the horizon $T_f$, thereby providing more accurate and context-aware decision support for autonomous driving.

\section{Semantic-Driven Agentic AI Empowered Framework with Neighboring Vehicle} \label{sec:5}
In the V2V communication scenario, this paper further proposes a semantic-driven Agentic AI empowered framework in which each vehicle predicts its future trajectory over the next $T_f$ time steps by leveraging its own historical motion data and the predicted future trajectories of neighboring vehicles. To alleviate communication latency and reduce the cognitive burden of processing high-dimensional raw data, semantic communication is incorporated into the V2V exchange. By transmitting semantically compressed and context-aware trajectory information, vehicles achieve a deeper understanding of their local traffic environment while lowering communication overhead, thereby enhancing collaborative perception.

\begin{figure*}[htbp]       
     \centering
        \includegraphics[width=\textwidth]{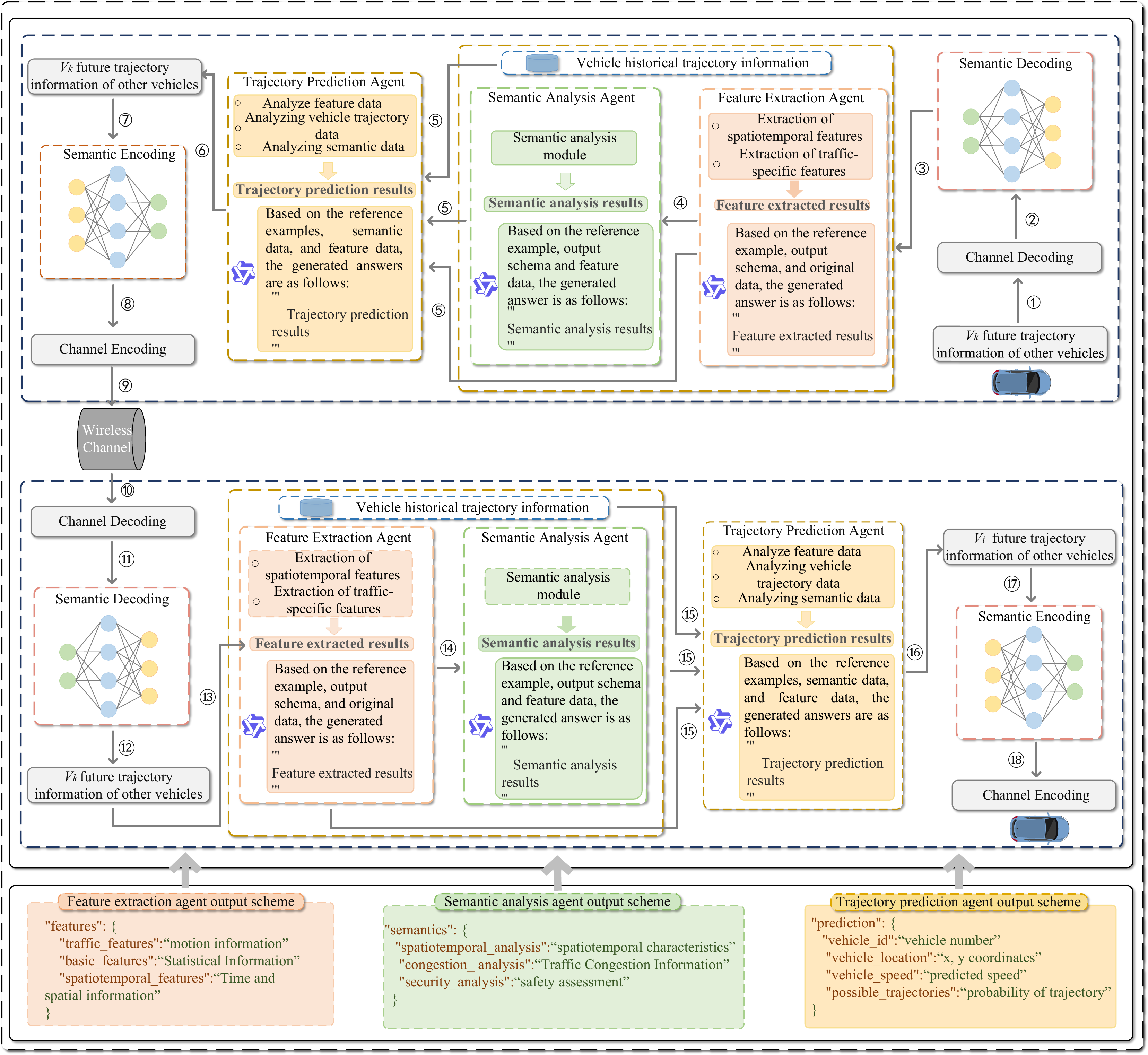}
        \caption{V2V network diagram. Neighboring vehicles transmit their future trajectory information via semantic communication. The target vehicle performs feature extraction and semantic analysis on its own trajectory data and predicts its future trajectory by jointly considering its own information and the received future trajectories of neighboring vehicles.}
        \label{fig:mt6}
\end{figure*}

The corresponding network architecture is shown in Fig.~\ref{fig:mt6}. Each vehicle is equipped with local vehicle information, a channel encoder/decoder, a semantic encoder/decoder, a feature extraction agent, a semantic analysis agent, and a trajectory prediction agent. Vehicles broadcast their predicted future trajectories to neighboring vehicles via semantic communication, enabling efficient information sharing and cooperative prediction.

\subsection{Feature Extraction Agent in V2V}
To obtain a more comprehensive representation of a vehicle’s motion behavior, the feature extraction agent in V2V operates on the vehicle’s own historical trajectory data. It filters and extracts key features, including spatiotemporal dynamics (e.g., velocity and acceleration profiles), statistical descriptors of trajectory distributions, and semantic features reflecting driving intent within the road environment.

To enhance extraction accuracy and consistency, prompt engineering is employed. By designing structured, semantically explicit prompts, the feature extraction task is reformulated as a semantic querying and reasoning process over trajectory data. This design guides the LLM to accurately interpret motion semantics and output feature representations with clear physical meaning and predictive relevance. These high-quality feature vectors serve as informative inputs for both the semantic analysis agent and the trajectory prediction agent.

\subsection{Semantic Analysis Agent in V2V}
To enable vehicles to thoroughly interpret their motion patterns and local traffic context, a semantic analysis agent is also deployed in the V2V framework. This agent receives multidimensional feature representations from the feature extraction stage and performs semantic reasoning and contextual analysis to infer richer environment-aware information from the viewpoint of the individual vehicle.

The inferred information includes congestion levels and traffic efficiency, traffic-flow characteristics across different time periods, periodic behavioral patterns, and segment-specific traffic properties. Similar to the V2I case, prompt-based guidance is applied to regulate the semantic analysis process. Carefully designed task-oriented prompts assist the LLM in integrating scene knowledge with trajectory features for joint reasoning. The prompt design explicitly accounts for traffic complexity and diversity, and enforces consistency with physical constraints and realistic driving behaviors, thereby ensuring that the derived semantic insights align with actual traffic dynamics.

\subsection{Trajectory Prediction Agent in V2V}
Since the future trajectories of neighboring vehicles strongly influence the motion decisions of the target vehicle, the proposed framework leverages V2V semantic communication to exchange predicted trajectories among vehicles. For trajectory prediction, the target vehicle aggregates its own feature extraction results, semantic analysis outputs, and the received future trajectory information of neighboring vehicles. By fusing these multi-source inputs, the trajectory prediction agent is able to conduct multidimensional reasoning on the target vehicle’s future motion, thereby improving prediction accuracy and robustness.

To further enhance prediction performance, prompt engineering is again employed. Structured prompts are constructed to guide the LLM to jointly process historical trajectories, local features, semantic analysis results, and neighboring vehicles’ future trajectories as a unified reasoning task. This design enables the trajectory prediction agent to more effectively capture interactive patterns and inter-vehicle dependencies.

\subsection{Semantic Communication}
In the V2V setting, vehicles exchange prediction-related information through semantic communication. Neighboring vehicles first transmit their predicted trajectories to the target vehicle using semantic encoders and channel encoders. Upon reception, the target vehicle applies channel decoding followed by semantic decoding to reconstruct trajectory data with explicit semantic meaning. It then performs feature extraction and semantic analysis on its own information, and finally predicts its future trajectory by combining these local results with the reconstructed trajectories from neighboring vehicles. Once a new prediction is generated, the target vehicle, in turn, broadcasts this prediction to neighbors via semantic communication. The underlying semantic communication system follows the same encoder–channel–decoder design as in the V2I case.

The V2V semantic communication process is described as follows:
\begin{itemize}
    \item[(1)] Semantic Encoding: Let the original vector of the emission be $x_{pr}$. First, semantic features are extracted through the semantic encoder $F_{se} (\cdot)$. The original vector is mapped to the feature vector $s_{pr}\in \mathbb{S}^k$. This process can be expressed as
    \begin{equation}
    s_{pr}=F_{se}(x_{pr},c_{pr},\alpha),  \label{eq:eq17}
    \end{equation}

    \noindent where, $F_{se} (\cdot)$ is a neural network, the $k$-dimensional vector $s_{pr}$ output from $F_{se} (\cdot)$ is the trajectory information of the feature extraction result, $\alpha$ is the parameter of the semantic encoder $F_{se} (\cdot)$, $c_{pr}$ is the channel bandwidth ratio of the channel, which is the ratio of the feature vector dimension $k$ of the original data dimension $m_{pr}$, which can be expressed as
    \begin{equation}
    c_{pr}=\frac{k}{m_{pr}}. \label{eq:eq18}
    \end{equation}

    \item[(2)]  Wireless Communication: When transmitting over a fading wireless channel, the composite vector $s_{pr}$ is subject to transmission losses including distortion and noise. This transmission process can be modeled as:
    \begin{equation}
    y_{pr}=H \cdot s_{pr} + N, \label{eq:eq19}
    \end{equation}

    \noindent where $y_{pr}$ denotes the received complex vector, $H$ represents the channel gain between the transmitter and receiver, and $N$ signifies additive white Gaussian noise (AWGN).

    \item[(3)] Semantic Decoding: Upon receiving vector $y_{pr}$, the semantic decoder for vehicle $v_k$ reconstructs it. This process can be represented as follows:  
    \begin{equation}
    \hat{x}_{pr} = F_{sd}(y_{pr}, c_{pr}, \beta), \label{eq:eq20}
    \end{equation}

    \noindent where, $F_{sd} (\cdot)$ is a neural network. The vector $y_{pr}$ output from $F_{sd} (\cdot)$ represents the decoded and reconstructed trajectory information. $\beta$ denotes the parameters of the semantic decoder $F_{sd} (\cdot)$.
\end{itemize}

\section{Experimental Results}
\label{sec:6}
\subsection{Dataset}
This paper evaluates the proposed method on the US-101 subset of the NGSIM dataset \cite{coifman2017critical}. The data were collected using high-resolution cameras installed along the US-101 freeway in Los Angeles, California, capturing high-frequency traffic trajectories over multiple time periods, including peak and off-peak hours. The dataset provides $45$-minute trajectory recordings under light, moderate, and heavy traffic conditions. Following \cite{deo2018convolutional}, vehicle trajectories are segmented into $8$-s clips, with the first $3$~s used as historical input and the subsequent $5$~s as the prediction horizon.

\subsection{Evaluation Indicators}
To measure the degree of discrepancy between predicted values and actual values, this paper employs the root mean square error (RMSE) for evaluation. The formula for calculating RMSE is as follows:
\begin{multline}
    RMSE = \\ \sqrt{\frac{1}{S} \sum_{s=1}^{S} {(\hat{x}_v^n(t_{T+s}) - x_v^n(t_{T+s}))}^2 + {(\hat{y}_v^n(t_{T+s}) - y_v^n(t_{T+s})})^2}. \label{eq:eq21}
\end{multline}

To measure the average linear distance deviation between predicted values and actual values, this paper employs the ADE for evaluation. The formula for calculating ADE is as follows:
\begin{multline}
    ADE = \\ \frac{1}{S} \sum_{s=1}^{S} \sqrt{  {(\hat{x}_v^n(t_{T+s}) - x_v^n(t_{T+s}))}^2 + {(\hat{y}_v^n(t_{T+s}) - y_v^n(t_{T+s})})^2}. \label{eq:eq22}
\end{multline}

To measure the straight-line distance deviation between the predicted trajectory point and the actual trajectory point at the very end of the prediction timeframe. This paper employs the FDE for evaluation, calculated as follows:
\begin{multline}
    FDE = \\ \sqrt{  {(\hat{x}_v^n(t_{T+S}) - x_v^n(t_{T+S}))}^2 + {(\hat{y}_v^n(t_{T+S}) - y_v^n(t_{T+S})})^2}. \label{eq:eq23}
\end{multline}.

\subsection{Experimental Analysis in V2I}
To assess the robustness of the proposed V2I scheme, we evaluate performance under different channel conditions with $\mathrm{SNR} \in \{0,10,20\}$~dB, corresponding to poor, moderate, and high-quality channels, respectively. FDE is used to assess the endpoint accuracy of single-trajectory prediction ($K=1$), whereas ADE and RMSE are used to evaluate multi-trajectory predictions ($K=5$ and $K=10$). The results are summarized in Table~\ref{tab:1}.

\begin{table}[htbp]
\centering
\caption{Performance under Different SNR Conditions in V2I}
\label{tab:1}
\begin{tabular}{cccccc}
\toprule
SNR & $K=1$ & \multicolumn{2}{c}{$K=5$} & \multicolumn{2}{c}{$K=10$} \\
\cmidrule(lr){2-2} \cmidrule(lr){3-4} \cmidrule(lr){5-6}
& FDE & ADE & RMSE & ADE & RMSE \\
\midrule
SNR=0~dB  & 4.67 & 3.33 & 3.84 & 1.64 & 1.85   \\
SNR=10~dB & 3.49 & 2.33 & 2.69 & 1.38 & 1.67   \\
SNR=20~dB & 2.49 & 1.15 & 1.39 & 0.83 & 0.92   \\
\bottomrule
\end{tabular}
\end{table}

As shown in Table~\ref{tab:1}, trajectory prediction accuracy degrades as SNR decreases. At $\mathrm{SNR}=0$~dB, strong channel noise causes severe semantic distortion and information loss, leading to the largest errors across all metrics. At $\mathrm{SNR}=10$~dB, improved channel conditions substantially reduce semantic distortion, enabling more accurate reconstruction of the environment and thus lower prediction errors. At $\mathrm{SNR}=20$~dB, semantic information is transmitted with minimal loss, allowing the model to infer trajectories from high-fidelity semantic features and achieving the best overall performance.

The multi-trajectory prediction strategy further improves robustness. As $K$ increases from $1$ to $10$, both ADE and RMSE decrease, indicating that generating multiple candidate trajectories better captures diverse driving intentions and increases the likelihood of covering the ground-truth trajectory. FDE remains higher than ADE across settings, reflecting the inherent difficulty and higher uncertainty associated with endpoint prediction.

To further verify the contribution of each module in V2I, we conduct ablation experiments at $\mathrm{SNR}=20$~dB under three configurations: (i) LLM only (no feature extraction, no semantic analysis), (ii) LLM-F (with feature extraction, without semantic analysis), and (iii) SemAgent (with both feature extraction and semantic analysis). The results are reported in Table~\ref{tab:2}.

\begin{table}[htbp]
\centering
\caption{Performance Comparison of Different Models at $\mathrm{SNR}=20$~dB in V2I}
\label{tab:2}
\begin{tabular}{cccccc}
\toprule
Method & \multicolumn{1}{c}{$K=1$} & \multicolumn{2}{c}{$K=5$} & \multicolumn{2}{c}{$K=10$} \\
\cmidrule(lr){2-2} \cmidrule(lr){3-4} \cmidrule(lr){5-6}
& FDE & ADE & RMSE & ADE & RMSE \\
\midrule
LLM (V2I)    & 4.33 & 3.20 & 3.51 & 2.07 & 2.39   \\
LLM-F (V2I) & 2.84 & 1.81 & 2.14 & 1.28 & 1.46   \\
SemAgent (V2I)   & 2.49 & 1.15 & 1.39 & 0.83 & 0.92   \\
\bottomrule
\end{tabular}
\end{table}

The LLM-only configuration yields the largest errors, indicating that relying solely on historical trajectories without auxiliary semantic information limits the model’s understanding of the traffic context. Incorporating feature extraction (LLM-F) substantially improves all metrics, confirming that structured environmental features are beneficial for trajectory prediction. The full SemAgent configuration achieves the best performance in all settings, demonstrating that semantic analysis effectively interprets and enriches the extracted features, thereby further enhancing prediction accuracy.

\begin{figure}[htbp]
    \centering
    \subfloat[\footnotesize ADE]{%
        \includegraphics[width=0.44\textwidth]{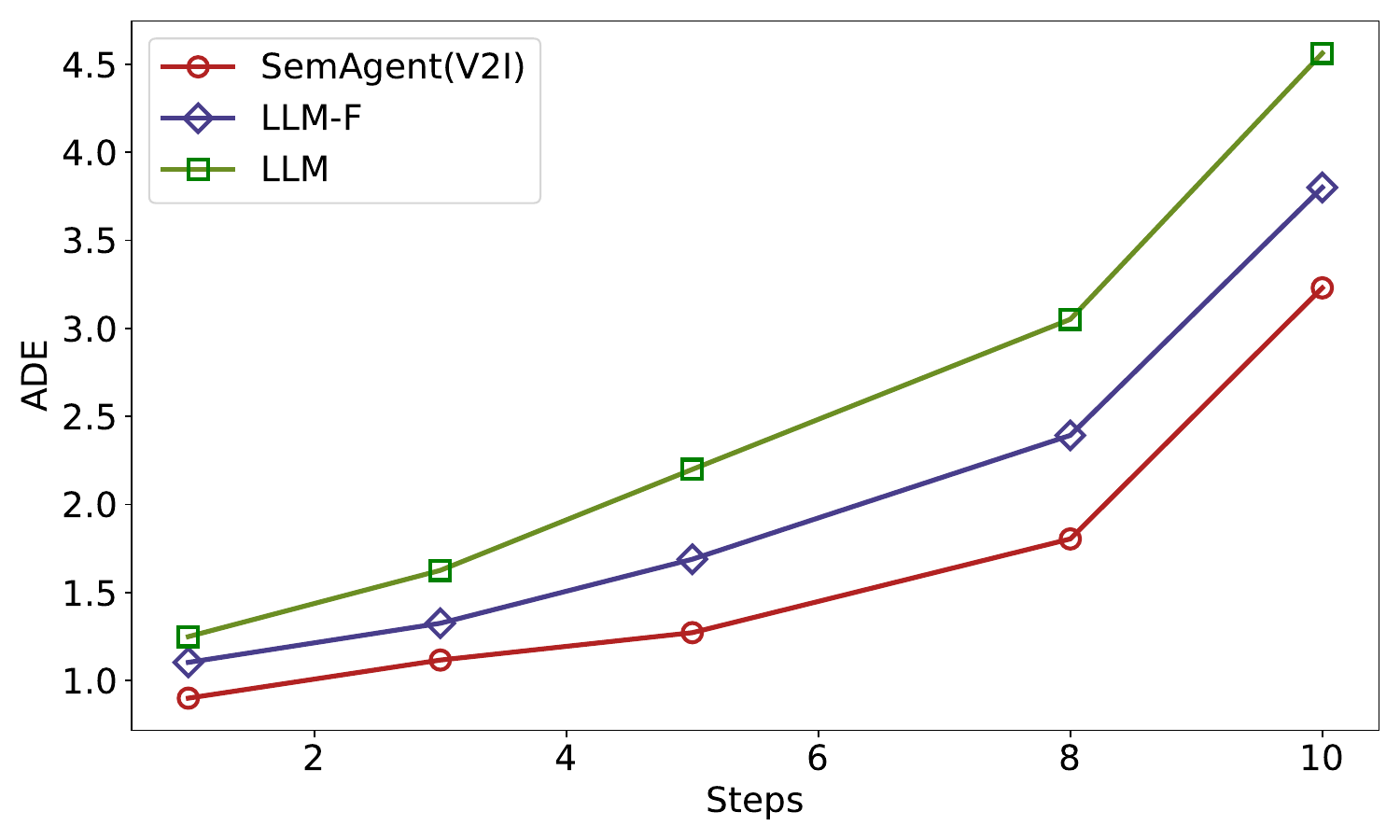} \label{fig:mt11}}\\
    \subfloat[\footnotesize FDE]{%
        \includegraphics[width=0.44\textwidth]{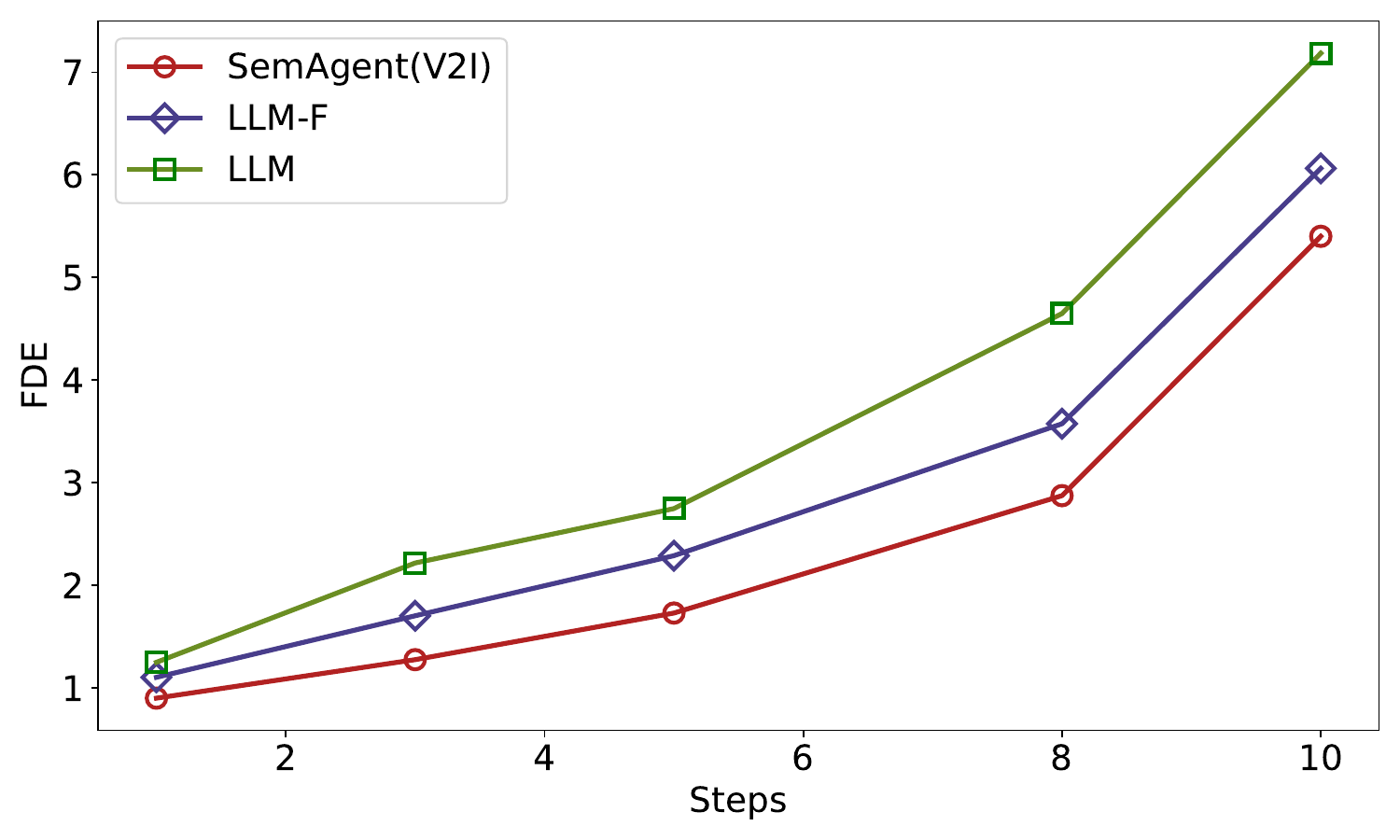} \label{fig:mt12}}\\
    \subfloat[\footnotesize RMSE]{%
        \includegraphics[width=0.44\textwidth]{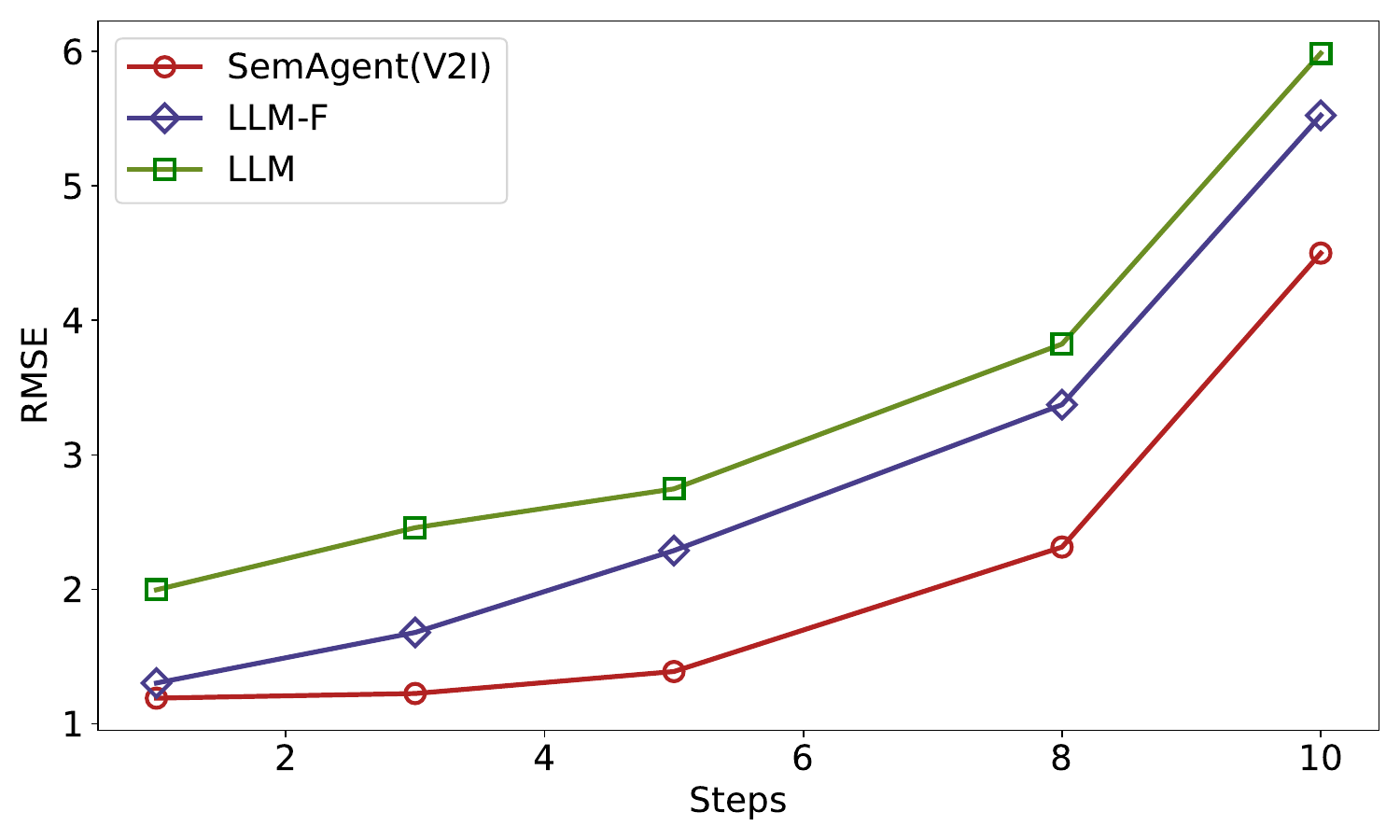} \label{fig:mt13}}
    \caption{Performance of different models in V2I for varying prediction horizons. The horizontal axis denotes the number of prediction steps, and the vertical axis denotes the corresponding metric value.}

    \label{fig:mt14}
\end{figure}

We also investigate the impact of prediction horizon length under $\mathrm{SNR}=20$~dB and $K=1$. The results, shown in Figs.~\ref{fig:mt11}–\ref{fig:mt13}, indicate that ADE, FDE, and RMSE gradually increase as the number of prediction steps grows. This behavior reflects the intrinsic uncertainty of long-horizon trajectory prediction: as the prediction window extends, the number of plausible future paths increases, and the model’s estimates tend to converge toward an averaged or most-likely trajectory, which deviates more from the actual realized path. Across all horizons, the proposed model consistently achieves lower errors than baselines. By integrating contextual environmental information and injecting deep and semantic features via semantic communication and multi-agent collaboration, the model better captures complex environmental dynamics, mitigates long-term error accumulation, and improves overall prediction accuracy.

\subsection{Experimental Analysis in V2V}
To evaluate the proposed framework in V2V communication, we apply the same channel settings and examine the vehicle-side prediction module. The results are presented in Table~\ref{tab:3}.

\begin{table}[htbp]
\centering
\caption{Performance of the model under different SNR conditions in V2V}
\label{tab:3}
\begin{tabular}{ccccccccc}
\toprule
SNR & \multicolumn{1}{c}{$K=1$} & \multicolumn{2}{c}{$K=5$} & \multicolumn{2}{c}{$K=10$} \\
\cmidrule(lr){2-2} \cmidrule(lr){3-4} \cmidrule(lr){5-6}
 & FDE & ADE & RMSE & ADE & RMSE \\
\midrule
SNR=0~dB  & 4.39 & 3.36 & 4.39 & 1.89 & 2.32   \\
SNR=10~dB & 4.02 & 2.79 & 3.71 & 1.44 & 1.54   \\
SNR=20~dB & 2.88 & 1.39 & 1.78 & 0.86 & 1.01   \\
\bottomrule
\end{tabular}
\end{table}
Similar to the V2I case, prediction accuracy in V2V improves monotonically with SNR. At $\mathrm{SNR}=0$~dB, strong interference severely degrades the quality of exchanged semantic information, weakening cooperative perception and yielding large errors. At $\mathrm{SNR}=10$~dB, performance lies between the $0$~dB and $20$~dB cases, confirming a positive correlation between channel quality and trajectory prediction accuracy. At $\mathrm{SNR}=20$~dB, vehicles can reliably share high-fidelity trajectory features and semantic context, achieving the best performance. Increasing $K$ again reduces ADE and RMSE by covering multiple plausible driving intentions, although FDE remains relatively higher due to endpoint uncertainty.

To further examine the contribution of each component in V2V, we perform ablation experiments at $\mathrm{SNR}=20$~dB under several configurations: (i) LLM (no feature extraction, no semantic analysis, no neighboring trajectory information), (ii) LLM-F (with feature extraction only), (iii) LLM-P (with neighboring trajectory information only), (iv) LLM-FS (with feature extraction and semantic analysis), (v) LLM-FS/FP variant, and (vi) SemAgent (full framework with feature extraction, semantic analysis, and neighboring trajectory information). The results are given in Table~\ref{tab:4}.

\begin{table}[htbp]
\centering
\caption{Performance comparison of different models when $SNR=20$ in V2V}
\label{tab:4}
\begin{tabular}{ccccccccc}
\toprule
Method & \multicolumn{1}{c}{K=1} & \multicolumn{2}{c}{K=5} & \multicolumn{2}{c}{K=10} \\
\cmidrule(lr){2-2} \cmidrule(lr){3-4} \cmidrule(lr){5-6}
 & FDE & ADE & RMSE & ADE & RMSE \\
\midrule
LLM (V2V)  & 5.39 & 2.84 & 3.39 & 1.89 & 2.26   \\
LLM-F (V2V) & 2.81 & 1.96 & 2.66 & 1.63 & 1.77   \\
LLM-P (V2V) & 3.15 & 1.95 & 2.65 & 1.51 & 2.21   \\
LLM-FS (V2V) & 2.53 & 1.49 & 1.71 & 1.07 & 1.32   \\
LLM-FS (V2V) & 2.52 & 1.41 & 1.72 & 1.25 & 1.43   \\
SemAgent (V2V) & 2.88 & 1.39 & 1.78 & 0.86 & 1.01   \\
\bottomrule
\end{tabular}
\end{table}
The baseline LLM configuration yields the largest errors for all $K$ values, indicating that neglecting feature extraction, semantic analysis, and neighboring information severely limits predictive performance. Incorporating either feature extraction (LLM-F) or neighboring trajectory information (LLM-P) alone significantly reduces all metrics, confirming that both structured features and cooperative predictions are beneficial. Combining feature extraction with semantic analysis (LLM-FS) yields further gains, illustrating that semantic reasoning can effectively exploit extracted features to improve prediction quality. When neighboring predictions are also considered, the framework better captures vehicle–vehicle interactions and further enhances accuracy. Overall, the full SemAgent-based V2V setting achieves consistently strong performance, validating the effectiveness of the proposed holistic design for multi-agent trajectory prediction.

\begin{figure}[htb]
    \centering
    \subfloat[\footnotesize ADE]{%
        \includegraphics[width=0.44\textwidth]{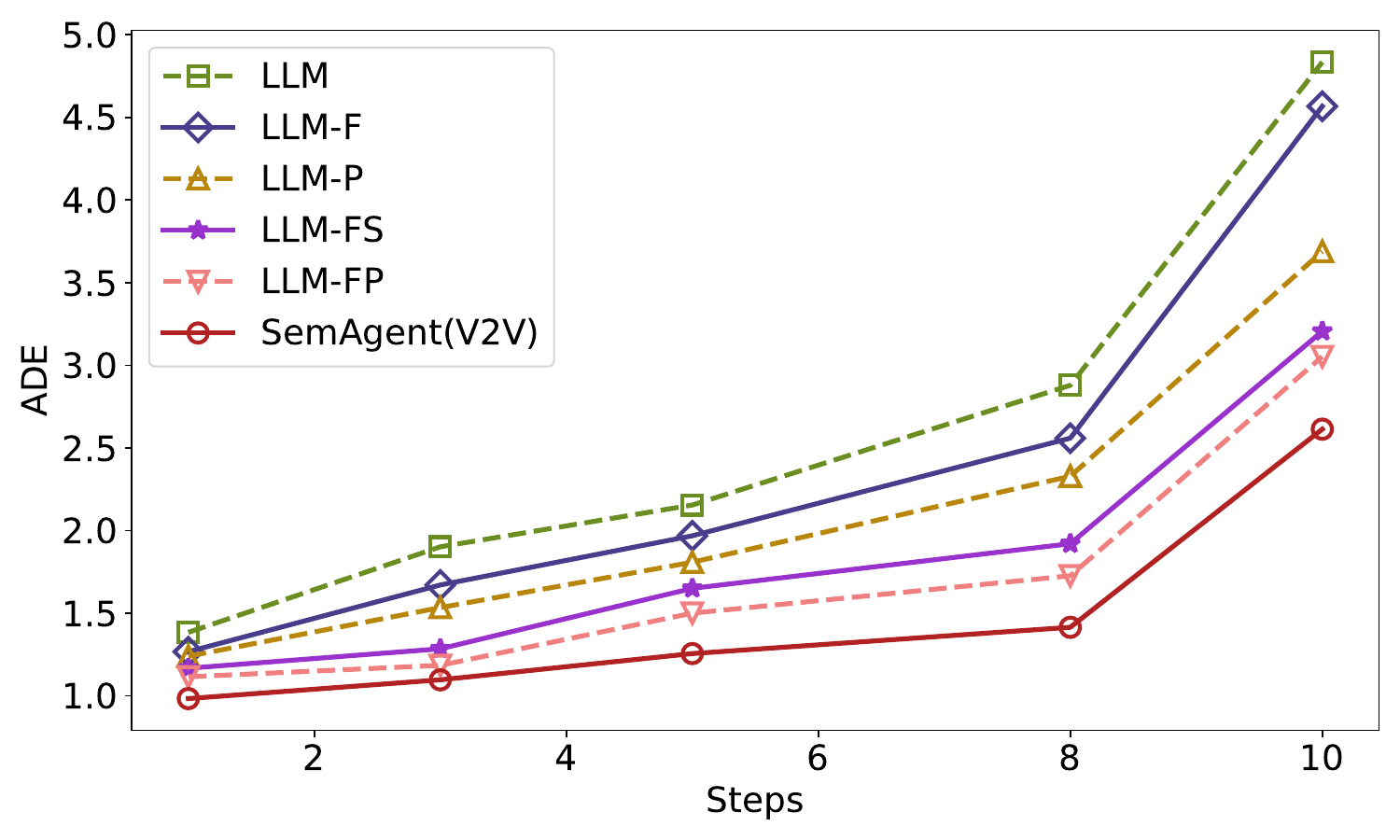} \label{fig:mt15}}\\
    \subfloat[\footnotesize FDE]{%
        \includegraphics[width=0.44\textwidth]{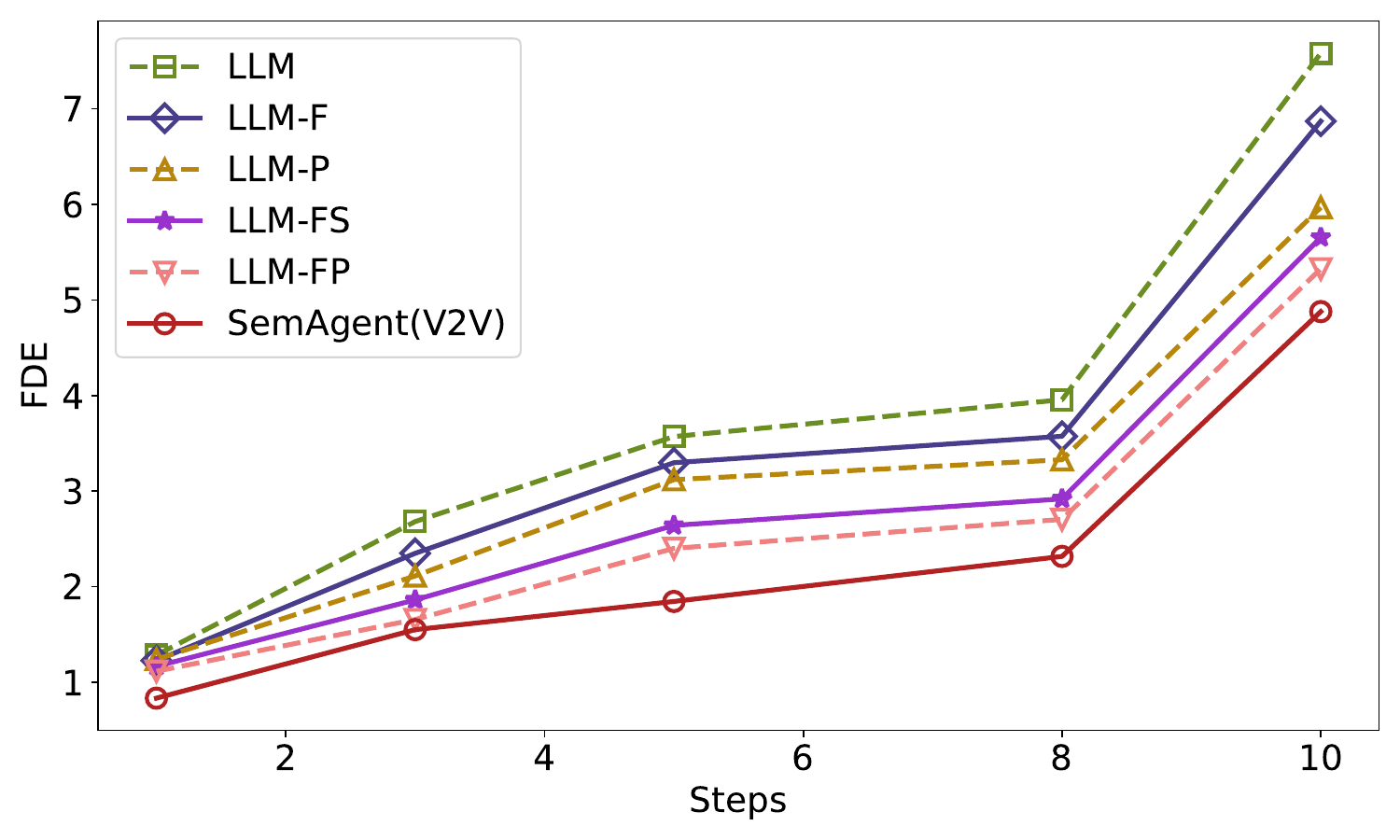} \label{fig:mt16}}\\
    \subfloat[\footnotesize RMSE]{%
        \includegraphics[width=0.44\textwidth]{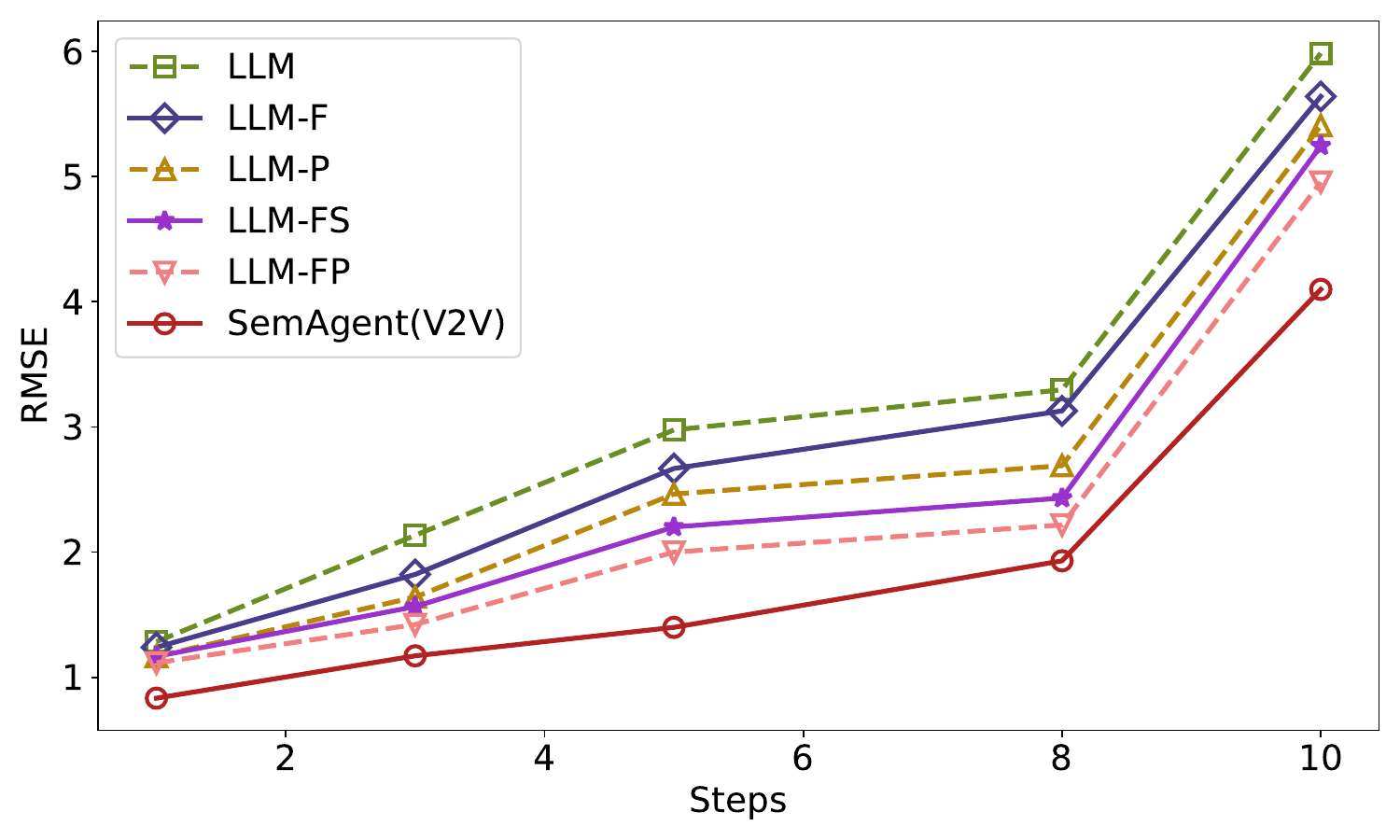} \label{fig:mt17}
    }
    \caption{Performance of different models in V2V for varying prediction horizons. The horizontal axis denotes the number of prediction steps, and the vertical axis denotes the corresponding metric value.}
    \label{fig:mt18}
\end{figure}

We also examine the impact of prediction horizon length in V2V under $\mathrm{SNR}=20$~dB and $K=1$. The results in Figs.~\ref{fig:mt15}–\ref{fig:mt17} show that ADE, FDE, and RMSE increase with the number of prediction steps. In V2V scenarios, each vehicle’s future trajectory is influenced not only by its own dynamics but also by the complex, time-varying behaviors of neighboring vehicles. As the prediction horizon grows, interactions are amplified and the joint behavior space expands combinatorially, making accurate long-term prediction increasingly challenging. Consequently, model estimates tend to converge to statistically plausible average behaviors, leading to accumulated errors.

Despite this, the proposed method achieves the best performance across all horizons. By explicitly modeling inter-vehicle interactions and exploiting future trajectory information from neighboring vehicles via semantic communication, the framework effectively mitigates conflicts between agents and improves the fidelity of predicted trajectories.

\section{Conclusion}
\label{sec:7}
This paper presented a vehicle trajectory prediction framework that integrates semantic communication with Agentic AI to enhance perception and reasoning in complex traffic environments. Within the V2X context, the framework is instantiated for both V2I and V2V communication modes through multiple coordinated agents with distinct roles.

In the V2I setting, the RSU employs a feature extraction agent and a semantic analysis agent to derive and interpret global traffic environment information. The resulting semantic representations are efficiently transmitted to vehicles via semantic communication. After decoding, vehicles obtain a semantically enriched understanding of their surroundings, significantly improving future trajectory prediction accuracy. In the V2V setting, target vehicles receive predicted future trajectories from neighboring vehicles through semantic communication and, in parallel, apply their own feature extraction and trajectory prediction agents to process historical trajectory data. By fusing local features, semantic analysis results, and neighboring vehicles’ predicted trajectories, the framework enables high-precision, context-aware trajectory prediction.

Extensive experiments on the NGSIM US-101 dataset demonstrate that the proposed approach consistently outperforms baseline methods across various channel conditions and prediction settings, confirming the effectiveness of combining semantic communication and Agentic AI for multi-agent trajectory prediction in intelligent transportation systems.


\bibliographystyle{IEEEtran} 
\bibliography{cas-refs}

\newpage

\end{document}